# Learning cross-lingual phonological and orthagraphic adaptations: a case study in improving neural machine translation between low-resource languages

*Saurav Jha*[1]*, Akhilesh Sudhakar*[2]*, and Anil Kumar Singh*[2]
[1] MNNIT Allahabad, Prayagraj, India
[2] IIT (BHU), Varanasi, India

## Abstract

Out-of-vocabulary (OOV) words can pose serious challenges for machine translation (MT) tasks, and in particular, for low-resource language (LRL) pairs, i.e., language pairs for which few or no parallel corpora exist. Our work adapts variants of seq2seq models to perform *transduction* of such words from Hindi to Bhojpuri (an LRL instance), learning from a set of cognate pairs built from a bilingual dictionary of Hindi – Bhojpuri words. We demonstrate that our models can be effectively used for language pairs that have limited parallel corpora; our models work at the character level to grasp phonetic and orthographic similarities across multiple types of word adaptations, whether synchronic or diachronic, loan words or cognates. We describe the training aspects of several character level NMT systems that we adapted to this task and characterize their typical errors. Our method improves BLEU score by 6.3 on the Hindi-to-Bhojpuri translation task. Further, we show that such transductions can generalize well to other languages by applying it successfully to Hindi – Bangla cognate pairs. Our work can be seen as an important step in the process of: (i) resolving the OOV words problem arising in MT tasks ; (ii) creating effective parallel corpora for resource constrained languages

*Keywords: Neural machine translation, Hindi - Bhojpuri, word transduction, low resource language, attention model*

; and (iii) leveraging the enhanced semantic knowledge captured by word-level embeddings to perform character-level tasks.

1                Introduction

With recent advances in the field of machine translation (MT) – and in neural machine translation (NMT) in particular – there has been an increasing need to shift focus to out-of-vocabulary (OOV) words (Wu *et al.* 2016; Sennrich *et al.* 2016). In the case of low resource languages (LRLs), which lack linguistic resources such as parallel corpora, most words are OOV words; this is problematic. Current data-intensive translation systems work poorly with OOV words for such languages, purely because of a severe lack of resources. Hence, for such languages, it becomes necessary to deal with OOV words in specific ways, outside the ambit of general-purpose NMT systems. Even in the case of resource-rich languages, methods to deal with OOV words are still being actively researched (Pham *et al.* 2018; Luong and Manning 2016).

Many approaches (as elaborated upon in Section 2) have been investigated to deal with the OOV problem. In this paper, we use the method of 'transduction', learned from a dictionary of cognate word pairs in Hindi and Bhojpuri. The fundamental guiding principle of our approach is the fact that Bhojpuri and Hindi are closely related languages, and hence have a good amount of vocabulary overlap, while sharing orthographic and phonetic traits. These two languages have common ancestors, and both of them employ the orthographically shallow alpha-syllabic Devanagari script. Tracing the origin of a considerable portion of the modern Bhojpuri vocabulary could weakly suggest that many of Hindi words got adapted to the local Bhojpuri phonology with the passage of time. This is a well-known phenomenon and it can be observed in other pairs of closely related languages (Macedonian – Bulgarian; Spanish – Catalan; Turkish – Crimean Tatar; Czech – Slovak; Irish – Scottish Gaelic) which share a close ancestor within the language family they belong to.

The Indian linguistic space, as reported by Grierson (1928), has 179 independent languages and 544 dialects. Similarly, the survey of Mahapatra *et al.* (1989) demonstrates that there are at least 50 Indian languages in which writing and publishing are done in substantial quantity. However, a majority of these languages lack proper linguistic resources. Hindi – being the *lingua franca* of the 'Hindi belt' (most parts of the north) of India – is a commonly spoken language in more than 10 states and has (according to one of the views[1]) seven major closely related languages, often called 'dialects' or 'sub-languages', (namely, *Awadhi, Bagheli, Bhojpuri, Bundeli, Haryanvi, Kanauji* and *Khari Boli*) (Mishra and Bali 2011). Despite it having 33 million native speakers in India and over 6 million native speakers outside India[2], Bhojpuri still suffers from the lack of language resources. So far, very little work has been done towards developing language resources (Singh and Jha 2015), resulting in scarcity of resources such as a Bhojpuri lexical database or parallel corpus that could have made state-of-the-art machine translation (MT) systems accessible to this language.

Due to the lack of such resources, traditional phrase based machine translation (PBMT) approach (Chiang 2005) or NMT (Bahdanau *et al.* 2014) for Bhojpuri becomes infeasible as it requires massive parallel corpora. In their recent work, Sharma and Singh (2017) introduce a 'word transduction' approach to deal with the presence of unknown (or out of vocabulary) words for MT systems involving such resource-scarce languages. The concept of word transduction is somewhat similar to Hajic (2000), where the authors suggest that the use of word-for-word translation for very closely related languages provides a good solution for MT of such language pairs.

Furthermore, for the task of language translation, it is necessary to take into account the fact that not all languages possess the same morphological features. For example, Finnish has more than 2000 possible inflected noun forms;(Ekman and Järvelin 2007) Hindi and Bhojpuri have more than 40 inflectional forms (Singh and Sarma 2010); while English has a mere 7 – 8 (these numbers indicate the different possible valid combinations of morphological tags that nouns can possess). Therefore, a good MT system designed for such morphologically rich languages must be intricate enough to deal with their diverse inflectional morphology. In order to address this issue, we adapt character-level NMT systems to our task in order to exploit morphological information encoded in inter-character interactions and intra-word patterns. As observed by Nakov and Tiedemann (2012): "character-level models combine the generality of character-by-character transliteration and lexical mappings of larger units that could possibly refer to morphemes, words or phrases, as well as to various combinations thereof" (Nakov and Tiedemann (2012)). We also introduce a novel pre-trained character-level embedding (Bojanowski *et al.* 2017) for Devanagari alphabets derived from the 300 dimensional Hindi fastText embeddings.[3]

---

[1] There is no consensus about the meaning of the word 'Hindi' and so different scholars have different views. For example, some other sub-languages like Rajasthani, Maithili and Magahi are also often included in the Hindi spectrum. However, the usual meaning of the word 'Hindi' in literature refers to standard Hindi, whose base is Khari Boli and which is an official language of India.

[2] http://www.censusindia.gov.in/Census_Data_2001/Census_Data_Online/Language/data_on_language.aspx

As regards the phonetic considerations of transduction, we make use of the fact that Hindi and Bhojpuri have a phonetic writing system, meaning there is an almost one-to-one mapping between phonemes (pronunciation) and graphemes (transcription). This is due to the fact that they both derive from common ancestor languages such as Prakrit and then Apabhramsha (Choudhury 2003). Hence, it suffices to work in either one of the spaces – orthographic or phonetic, and we choose the orthographic space since it does away with the need to convert the graphemes of text to and from phonemes.

Although Bhojpuri phonology is close to that of Hindi, it is not the same. There are notable differences between the two. While Hindi has a symmetrical ten vowel system, Bhojpuri has six vowel phonemes and ten vocoids. Similarly, Hindi has 37 consonants (including those inherited from earlier Indo-Aryan and those from loan words), whereas Bhojpuri has 31 consonant phonemes and 34 contoids. As is usual with any pair of languages, there are many phoneme sequences which are allowed in Hindi, but not found in Bhojpuri and vice-versa. This will be evident from the examples given in the paper later. More details about the Bhojpuri phonology are available in the article by Trammell (1971).

---

[3] https://github.com/facebookresearch/fastText/blob/master/docs/pretrained-vectors.md

## 1.1 *A note on Roman representations and English glosses of Hindi*

Throughout this paper, we have used the WX notation (Gupta *et al.* 2010) to represent (in a transliteration-like fashion) Hindi and Bhojpuri characters in English, for the benefit of readers who are not familiar with the Devanagari script. A ready reference table of the WX notation can be found in its Wikipedia page.[4] Every non-English word used in this paper is followed by its WX notation in square brackets ([]) and its italicized English gloss, in parenthesis (()).

## 1.2 *Transduction and translation*

Our usage of the word 'transduction' distinguishes it from translation, in that transduction is a task which is trained exclusively on cognates, and in that sense, the dataset it uses is a subset of the dataset that a translation system would use. **Cognates** are word pairs that not only have similar meaning but are also phonetically (and, in our case, orthographically) similar. The underlying observation that guides the usage of our proposed method of 'transduction' of OOV words as a possible substitute for their translation is as follows:

As stated earlier, Bhojpuri is a language closely related to Hindi. In the case of an OOV Hindi word (or any Hindi word for that matter), there is good chance that the Bhojpuri translation of the word is a cognate of the Hindi word adapted to the phonological and orthographic space of Bhojpuri due to the presence of borrowed words, common origins, geographic proximity, socio-linguistic proximity, etc. As a phonemic study of Hindi and Bundeli (Acharya 2015), mainly focusing on the prosodic features and the syllabic patterns of these two languages (unsurprisingly) concluded that the borrowing of words from Hindi to Bundeli generally follows certain (phonological) rules. For instance if a word in Hindi begins with the character [ya], it is replaced by character [ja] in its Bundeli equivalent as [yajamaan] *(host)* becomes [jajamaan], [yamunaa *(name of a river)*] becomes [jamunaa], etc. We observe that a similar process happens for Hindi – Bhojpuri. This category of word pairs is our main motivation behind the work described in this paper.

Our model is agnostic to what sort of words are considered to be OOV (based on their unigram probabilities, their parts-of-speech

---

[4] https://en.wikipedia.org/wiki/WX_notation

(POS), or whether named entities, etc.) because the above assumptions hold uniformly across the language pair. Section 2 specifies some of the metrics that are used to identify OOV words in related work.

Further, the above assumption (of transduction improving overall translation performance) has been demonstrated to be valid in the case of many closely related language pairs, in a number of previous works. For instance, Kondrak *et al.* (2003) extracted a parallel list of cognate word pairs and re-appended them to the parallel list of all word translations, thereby increasing the training weights of cognate words. Giving added importance to these cognate words, "resulted in a 10% reduction of the word alignment error rate, from 17.6% to 15.8%, and a corresponding improvement in both precision and recall." (Kondrak *et al.* (2003)). Mann and Yarowsky (2001) used cognates to expand translation lexicons, Simard *et al.* (1993) to align sentences in a parallel translation corpora, and Al-Onaizan *et al.* (1999) used cognate information to improve statistical machine translation.

Finally, transduction induces less sparsity in the model as compared to translation, because the hypothesis space is restricted to only functions that map words to their possible cognates. For closely related languages, the added reduction in sparsity also comes from the fact that there are consistent variations between how a source word transduces to its cognate target word. Hence, transduction is a task that performs better with a small training set than translation would when using a similarly complex model. This reduced sparsity enables transduction models to perform well on OOV words.

The ensuing section provides background about NMT systems and the manner in which we have adapted them to our task.

### 1.3 *Neural machine translation*

Neural machine translation (NMT) has delivered promising results in large-scale translation tasks such as Chinese-to-English (Tu *et al.* 2017) and English-to-French (Chen and Wu 2017). Initially, NMTs were used as a sub-component of the PBMT system such as for generating the conditional probabilities of phrase pairs (Cho *et al.* (2014)), for generating [machine learning] features for the PBMT, or for re-ranking the *n*-best hypotheses produced by the system (Kalchbrenner and Blunsom (2013), Sutskever *et al.* (2014a)). Such combined systems produced state-of-the-art results. One most appealing feature of NMTs is that they are largely memory efficient. Unlike PBMT systems, an NMT system does not require keeping track of phrase pairs or language models. Additionally, the work of Bentivogli *et al.* (2016) pointed out that NMTs offer a range of other superior attributes including:

- generation of outputs that require considerably lower post-edit efforts
- better translation quality in terms of BLEU score, Translation Edit Rate, and performing well on longer sentences
- fewer and/or less-severe errors in terms of morphology and word order

Most NMT systems today make use of the encoder-decoder based approach (e.g. Forcada 1997; Cho *et al.* 2014; Kalchbrenner and Blunsom 2013; Sutskever *et al.* 2014a), which consists of two recurrent neural networks (or their variants). The first encodes a variable-length source token $x$ into a fixed length vector and the second decodes the vector into a variable-length target token $y$. NMT approaches were initially designed to work at the word-level and translate sentences. However, noting the encouraging results of adapting NMTs to character-level translation by Vilar *et al.* (2007), we adapt NMTs to our character-level transduction. Figure 1 shows the architecture of such an encoder-decoder based NMT system performing character-level transduction. The model is trained over a parallel corpus to learn the maximum conditional probability of $y$ given $x$, i.e., $argmax_y\ p(y \mid x)$. Once trained, the model can then be used to generate the corresponding transduction for a given source word.

However, the performance of NMT degrades largely in the case of longer length sequences (words, in our case) due to the vanishing gradient problem (Bengio *et al.* 1994) arising during the training of the underlying RNN. So far, the use of an attention mechanism, as stated by Bahdanau *et al.* (2014), Luong *et al.* (2015a), Vinyals *et al.* (2015) and Yang *et al.* (2016) has been the most plausible solution to the aforementioned problem for RNNs and its variants. The concept of 'attention' takes into account the fact that in the task of translation, different tokens in a sequence are differentially informative, with the information carried by them being highly context dependent. Thus, for predicting each corresponding token, the model looks at the current

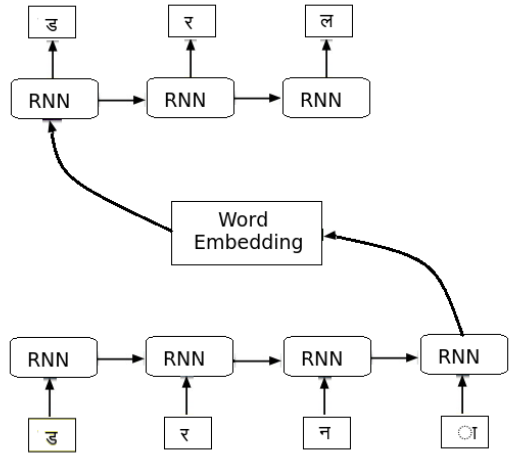

Figure 1: Encoder-decoder network architecture transducing the Hindi word [darnA] *(to be scared)* to Bhojpuri [darala]

context of the source token that is relevant to predicting the target token.

### 1.4 *Summary*

We adapt NMT models to perform 'transduction' of a Hindi word to a Bhojpuri word. These word-transduction models work with characters as the fundamental units. They are trained on Hindi – Bhojpuri cognate pairs. This task is important because it helps to solve the OOV problem in larger downstream tasks, the most prominent example of which is machine translation for low resource languages. To improve machine translation of Hindi to Bhojpuri, we first identify OOV words in Hindi texts and then use our model to transduce them to their Bhojpuri counterparts. All other words are translated, and not transduced. Using such separate treatment of OOV words, we obtain improvements upon BLEU score with respect to the originally translated texts. The section on Related Work (Section 2) elaborates previous approaches to transduction and the OOV problem.

## 2 Related work

A number of methods have been proposed to handle the OOV problems in machine translation of unknown or rare words. Luong *et al.* (2015a) simply use a shortlist of 30,000 most frequent words and map all other less frequent words to an UNK (unknown) token. Sutskever *et al.* (2014b) use a vocabulary of 80,000 words and achieve better performance. However, any UNK-based approach is problematic because in larger sentences, UNK tokens heavily degrade performance (Cho *et al.* (2014)). Jean *et al.* (2015) make model specific improvements, using a smaller batch for normalization and including only frequent words in the denominator of this normalization. They fall back to other translation and alignment models to replace UNK tokens.

Other approaches to handle OOV words include using a back-off dictionary look-up (Jean *et al.* (2015), Luong *et al.* (2015b)) but as observed by Sennrich *et al.* (2016), these techniques make impractical assumptions. One such assumption is a one-to-one source – target word correspondence, which is unwarranted. Further, some of these methods assume the existence of a parallel corpus of source-target word pairs, which is not always available in the case of low resource languages. Sennrich *et al.* (2016), in turn use a Byte-Pair Encoding for transduction, which is very similar to character-level encoding of sequences as strings of characters.

We also borrow ideas from previous approaches that have used cognates. Simard *et al.* (1993) use cognates to align sentences in a parallel corpus and report 97.6% accuracy on the alignments obtained, when compared to reference alignments. Mann and Yarowsky (2001) use cognates extracted based on edit distances for inducing translation lexicons based on transduction models. Scannell (2006) presents a detailed study on translation of a closely related language pair, Irish-Scottish Gaelic. They learn transfer rules based on alignment of cognate pairs, and use these rules to generate transductions on new words. They use a fine-grained cognate extraction method, by first editing Scottish words to 'seem like' Gaelic words, and then using edit string similarity on the new word pairs and choosing only close words with the additional constraint that both words in the pair should share a common English translation. However since we use linguistic experts to extract cognates from our dataset, we do not need to encode string similarity measures explicitly to extract cognates.

We borrow insights from character-level machine transliteration and translation models that have been proposed in the past, as transduction can be viewed as a variation of transliteration (which is, in turn, viewed as character-level translation in many works), working within the same script. Alternatively, it can also be thought of simply as a translation of 'true friend' cognates.

Vilar *et al.* (2007) work on transliteration at the character level (and translation at the word level) to build a combined system that shows increasing gains over just the word-level system, as the corpus size grows smaller. This is because the character-level transliteration takes into account the added morphological information such as base forms and affixes. Tiedemann experimented with different types of alignment methods and learning models, and showed that in each type of method, there exists at least one character-level model that performs better than word-level models (in the case of closely related language pairs). Denoual and Lepage (2006) also show merits of using characters as appropriate translations, and highlight issues with making assumptions about words being natural units for the task. Finch and Sumita (2009) view transliteration as a character-level machine translation, and use Phrase-Based SMT for bidirectionally encoding source sequences. They observe the lack of necessity to model phonetics of source or target language, due to the use of direct transformations. One point of difference between some of the related work on cognates and ours is that we do not perform context-sensitive transduction simply due to lack of annotated data that is context-sensitive.

3            Methodology

We run experiments on four different benchmark encoder-decoder networks, namely:

- a simple sequence-to-sequence model (Cho *et al.* (2014)) – abbreviated as **'seq2seq'**.
- the alignment method (Bahdanau *et al.* 2014) incorporated with the seq2seq model – abbreviated as **'AM'**.
- the Hierarchical Attention Network (Yang *et al.* 2016) incorporated with the seq2seq model – abbreviated as **'HAN'**.
- the Transformer Network (Vaswani *et al.* 2017) solely based on attention – abbreviated as **'TN'**.

In the basic seq2seq RNN encoder – decoder model (Cho *et al.* 2014) we incorporate a 'peek' at the context vector at every time step.

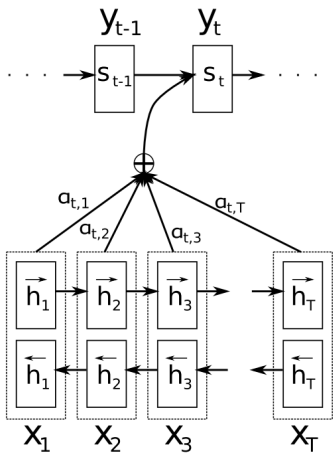

Figure 2:
Schematic of the alignment model adapted from Bahdanau *et al.* (2014)

However, the model performed poorly on this translation task, with the validation accuracy plateauing out at a low value early on in the training process. Section 3.1, Section 3.2 and Section 3.3 describe the models - AM, HAN and TN - in detail, respectively.

3.1    *Alignment model (AM)*

As shown in Figure 2, the alignment model (AM) proposed by Bahdanau *et al.* (2014) facilitates searching through the source sequence during the decoding phase using a unique context vector for each token. Specifically, given a translation $y_i$ and the source token **x**, the decoder decomposes the conditional probability over all the previously predicted tokens $(y_1, ..., y_{i-1})$ as:

$$p(y_i|y_1, ..., y_{i-1}, \mathbf{x}) = g(y_{i-1}, s_i, c_i) \quad (1)$$

where $s_i$ is the hidden state of the decoder model computed for time $i$, $c_i$ is the distinct context vector for each target token $y_i$ and $g$ is a non-linear function that outputs the probability of $y_i$ being the correct translation at time $i$.

In addition to the use of unique context vector for each decoding time step, the model takes into account that all the hidden states computed so far to contribute to the context vector $c_i$ with weight $\alpha$.

$$c_i = \sum_{j=1}^{T_x} \alpha_{ij} h_j \quad (2)$$

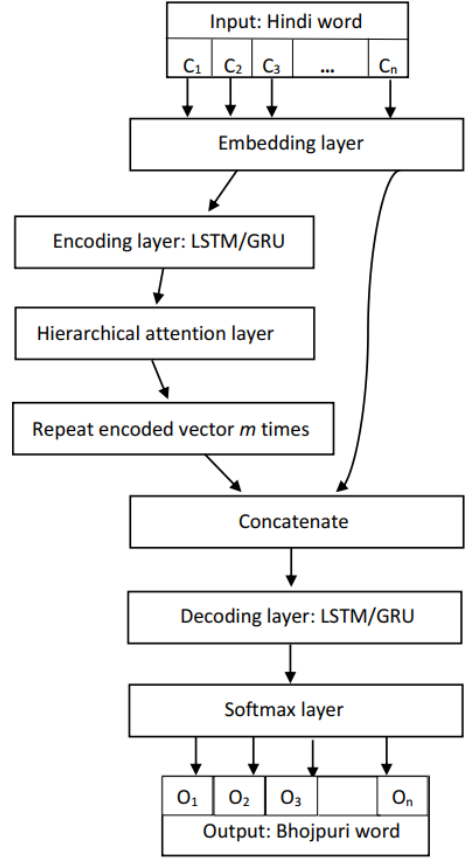

Figure 3: Encoder-decoder network architecture with HAN: $n$ and $m$ denote the number of characters in input and output words respectively

$\alpha$ thus serves as a normalized importance weight, measuring the degree of importance of the context tokens around position $j$ in predicting the translation of the current source token at the output position $i$ (see Bahdanau et al. (2014)). Figures 3 and 4 show the architecture of the encoder-decoder network incorporating the alignment-based attention decoder.

### 3.2 *Hierarchical attention network (HAN)*

Proposed by Yang *et al.* (2016), Hierarchical Attention Networks (HAN) exploit the hierarchical nature of documents (i.e., characters form words, words form sentences and sentences form a document) and are comprised of two levels of attention mechanisms (Bahdanau

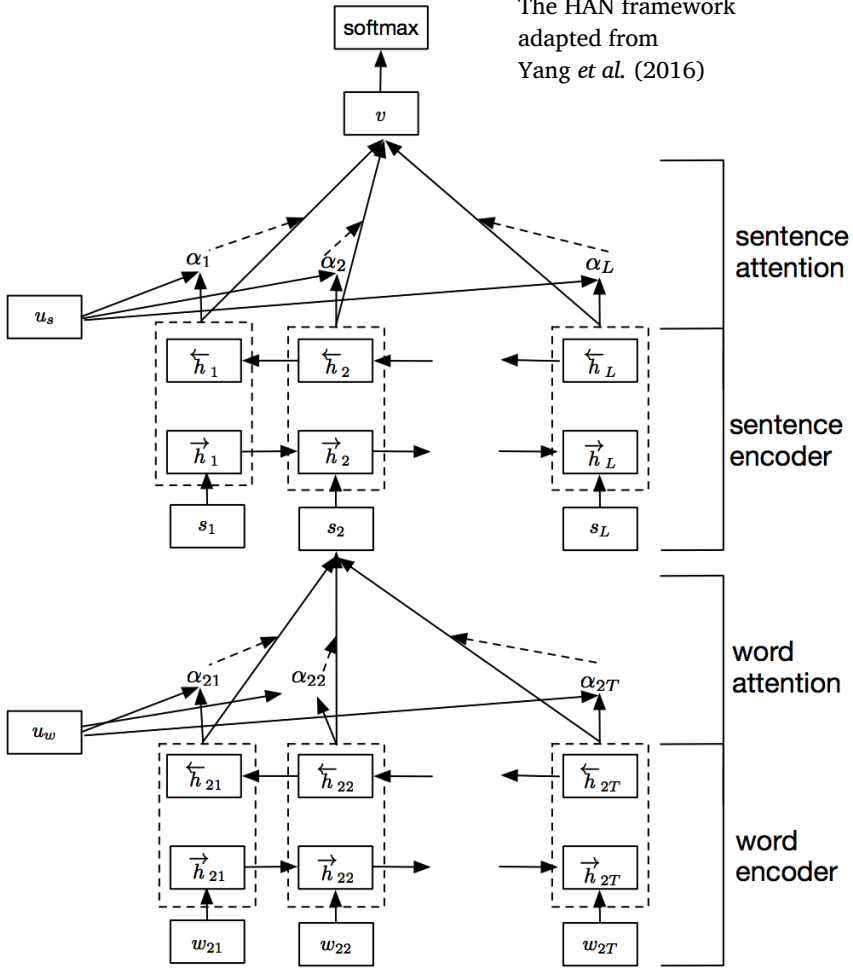

Figure 4: The HAN framework adapted from Yang *et al.* (2016)

*et al.* 2014; Lai *et al.* 2015)) – the first at the word level while the other at the sentence level. In our case, the former attention can be thought of being effective at the character level, while the latter at the word level, thus allowing the model (Figures 3 and 4) to discover the amount of attention required to be paid to the individual characters and words to form a character-level transduction.

### 3.3     *Self-attentional transformer network (TN)*

The Transformer Network (Vaswani *et al.* 2017) consists of an encoder made up of a stack of six identical layers. Each layer further consists of two sub-layers: *a multi-head self-attention* and a simple *position-wise* fully connected feed-forward network (FFN). The decoder too has a similar architecture except an additional sub-layer performing multi-head attention over the output of the encoder stack. Both the encoder and the decoder unit employ a residual connection (He *et al.* 2016) in between their respective sub-layers, followed by layer normalization (Ba *et al.* 2016).

As shown in Figure 5, the *positional embeddings* serve to make the representation at time step $i$ independent from the other time steps. The *multi-head attention* layer serves to replace recurrent dependencies by repeatedly applying self-attention over the same inputs using separate parameters (attention heads) followed by combining the results. This combination acts as an alternative to applying a single pass of attention with more parameters so that the model can easily learn to attend to different types of relevant information in parallel with each head. In other words, the decoder can now use multiple encoder-attention mechanisms in each of its layers resulting in a significantly faster training than architectures based on recurrent or convolutional layers. Inspired by the success of the Transformer in sequence generation tasks (such as achieving state-of-the-art results on both WMT2014 English-German and WMT2014 English-French translation tasks[5]), we use the Transformer Network in our task

---

[5] http://www.statmt.org/wmt14/

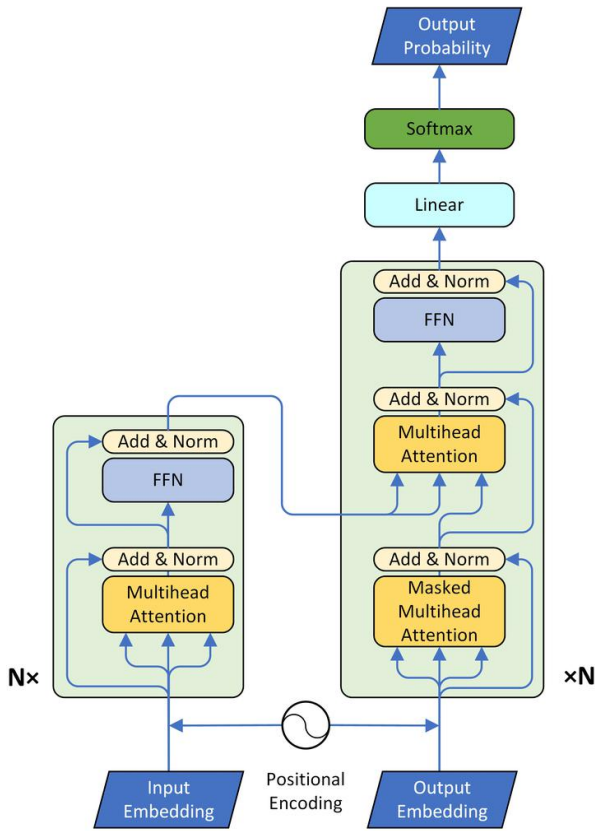

Figure 5: The transformer architecture as described in Vaswani *et al.* (2017) and adapted from Li *et al.* (2018)

## 4 Experiments

For the AM and HAN models, we consider various parameters while training, such as LSTM/GRU as encoding/decoding units, sequence chunking and batch sizes, optimization methods, regularization, and we report that there is a huge variance in the transduction performance depending on the combinations of the parameters used (Appendix A). As regards the TN model, due to computational and implementation limitations, we could not perform such an extensive hyperparameter search. Nevertheless, we take care of the basic parameter settings (as suggested in Popel and Bojar (2018)) by limiting our model to a single-GPU base, setting batch size to 512, maximum sequence length to the length of the longest token in the parallel corpus followed by a final averaging of the last 6 training checkpoints while leaving the learning rate and the number of warm up steps at their default values. Finally, the extracted character embeddings are used to train the AM and HAN models while the TN model is left void of these mainly because of its inherent dependency on segmenting the training tokens into semantically useful sub-tokens which are hard to be reproduced in varying experiments (Popel and Bojar (2018)), and thus cannot be easily assigned such embeddings. We compare the results of the TN model with the best results (among various hyperparameter settings) of the AM and HAN models.

The implementation of the Alignment model (AM) and the Hierarchical attention network (HAN) is based on Keras-2.0.6 (Chollet *et al.* 2015) while that for the Transformer Network (TN) is based on tensorflow-1.4.1 (Abadi *et al.* 2015) and tensor2tensor-1.4.3.[6] We carry out our experiments on x86_64 GNU/Linux with 8G memory, using one NVIDIA GeForce 840M with CUDA V8.0.61, and Python 3.5.2+.

### 4.1 *Dataset*

In order to be able to compare our results with the state of the art (SOTA) – described in Section 4.3 – we use the same dataset as by SOTA Sharma and Singh (2017), which was the state-of-the-art method at the time of writing. This dataset consists of 4220 Hindi – Bhojpuri word cognate pairs chosen from a pre-compiled lexicon of Hindi – Bhojpuri word translations. This dataset was compiled by three linguistic experts (who are native speakers of both Bhojpuri and Hindi) who came to consensus on the annotations. Cognate pairs were identified from this set using domain expertise by linguistic experts. This method of cognate extraction is in contrast to possibly sub-optimal rule-based and similarity-based approximations, such as those used by Mann and Yarowsky (2001). In summary, the training data has Hindi – Bhojpuri word pairs, each of which comprises a Hindi and a Bhojpuri word that have the same meaning, as well as similar pronunciations. We split the dataset into 3:1 ratio for training and testing our models. A validation split of 0.1 was further made on the train set comprising of 3165 word pairs. This test is held-out and is used for reporting only the final results. All hyperparameter tuning is done on the validation set. The validation set is not fixed and the training is cross-validated, with a new random validation-set being used at each iteration of tuning. Table 1 consolidates these statistics. We perform a random shuffle of the train and validation set prior to each training epoch.

---

[6] https://github.com/tensorflow/tensor2tensor

### 4.1.1     Semantic ambiguity when selecting cognates

It is worthwhile to mention that not all Hindi words in the training dataset have just one possible corresponding Bhojpuri cognate. Around 3.85% of Hindi words in the Hindi – Bhojpuri cognate pairs have more than one corresponding Bhojpuri cognate. The same applies the other way too, i.e., not all Bhojpuri words have just one possible corresponding Hindi cognate. Around 3.37% of Bhojpuri words in the Hindi – Bhojpuri cognate pairs have more than one corresponding Hindi cognate. When either a Hindi word or a Bhojpuri has multiple corresponding cognates, only two possible cases arise:

1. The multiple possible cognates have the same semantic sense and are different in only surface forms (spellings). For instance, the Bhojpuri word [haluVA] *(a sweet dish)* has two Hindi cognates - [halVA] and [halaVA], both meaning the same, but differing in surface forms due to the presence of the diacritic in one and absence in the other. The diacritic is a character explained in Section 6.1.1.

Table 1: Transduction corpus statistics

|  | Training set | Validation set | Test set |
|---|---|---|---|
| Total number of words | 2849 | 316 | 1055 |
| % of words of full dataset | 67.5% | 7.5% | 25% |

2. They are different inflected forms of the same root word. For instance, the Bhojpuri word [rAKala] *(kept, or to keep)* has two Hindi cognates - [raKA] *(kept)* and [raKanA] *(to keep)*, both of which are different inflected forms of the root word [raKa] *(keep)*.

Since multi-cognate words together form only a small percentage of the dataset, they are not accorded any special treatment, and the model learns from the multiple forms.

We plan to work more on the deeper change in ambiguity from source to target word in the case of 'true friend' cognates, particularly for the Hindi – Bhojpuri language pair.

4.2 *Evaluation measures*

We report the accuracy of each experiment using the BLEU score and Levenshtein distance-based string similarity (SS) measure, as in Equation 3.[7] After obtaining the optimum hyperparameter set for AM and HAN, we compare the word accuracy (WA, Equation 4) report defined by the percentage of correctly translated words for all the models including the SOTA. SS, WA and BLEU score formulae for two arbitrary strings 's1' and 's2' are given below. The averaged score across the validation/test set are reported in the tables in ensuing sections. We employ the character n-grams version of BLEU score (as used in Denoual and Lepage (2005)) as our work deals at the word-level, instead of document-level. Using this metric also alleviates the comparison with other state-of-art transduction methods since other popular metrics such as CHRF and TER correlate well with the character n-gram version of BLEU score (as observed by Popović (2015)).

$$SS(s1,s2) = (1 - \frac{Levenshtein\ Edit\ Distance(s1,s2)}{len(s1)+len(s2)}) * 100 \quad (3)$$

$$WA(s1,s2) = \{1\ if\ s1 == s2\ ;\ 0\ otherwise\} \quad (4)$$

---

[7] https://pypi.python.org/pypi/python-Levenshtein/0.12.0

4.3 *State of the art*

We consider the results of Sharma and Singh (2017) to be state of the art. To the best of our knowledge, their work is the only relevant one on Hindi – Bhojpuri transduction or even any form of OOV word handling technique for this language pair. While their work builds upon traditional PBMT approaches, they first convert lexical word representations into phoneme strings followed by the alignment of phonemes in these strings. The word is then segmented into phoneme chunks which thus facilitates the extraction of weighted rewrite rules for these chunks.

Since the work of Sharma and Singh (2017) extensively compares and contrasts their own work to the related work in transduction, we refrain from such an elaborate comparison, and suggest their work to the reader for more comparisons to other techniques. To the best of our knowledge, their work is the current state of the art, and we show improvements in performance over their results.

4.4 *Common aspects across models*

Our adaptations of each of the models (i.e., seq2seq, AM, HAN and TN) use Bidirectional LSTMs (BLSTM) as encoder-decoder units unless specified otherwise. A detailed analysis of hyperparameter tuning and training aspect for each model can be found in Appendix A. The appendix documents the experiments motivating our decisions on using LSTM vs GRU, sizes of encoding and decoding layers, number of layers, batching, optimization methods, regularization methods and pre-training embeddings for each of the three models. We find that these results may be useful for future work in morphology-related tasks. The results in each table in Appendix A show the epoch with least validation loss. We show the epoch (abbreviated as 'ep' in the tables) in order to keep track of the convergence speed and the risk of overfitting. The standard optimization method is (unless otherwise mentioned) Adam (Kingma and Ba 2014) with an initial learning rate of $10^{-3}$. The input to all these models is the Hindi word and the output is the transduced Bhojpuri word.

4.5 *Character embeddings*

Since the atomic token in our models is the character (and not the word), we explore two pre-training strategies (LM-LSTM-Embed and fT-Avg-Embed) to represent characters as dense vector embeddings. These embeddings are used only as pre-trainings, and the transduction models are allowed to update these weights during training, i.e., they are not frozen. The followings sections describe methods used to create these embeddings.

4.5.1 Creating embeddings using LSTM

Taking cues from Sundermeyer *et al.* (2012), we use a simple one-layered Bi-LSTM of hidden dimension 75 and a dropout of 0.5 to train a character-level language model for Hindi. Working at the character-level, we formulate the problem as a character prediction task – given a sequence of 29 characters, predict the next one, i.e., the $30^{th}$ character. We found 30 to be the optimum window size after varying window sizes from 5 to 40, keeping in mind memory requirements as well as perplexity. We randomly sample 3000 Hindi Wikipedia articles as a train set (with a validation split of 0.2) and another random 600 articles as a held-out test set. With a train set size of over 19M characters, a vocabulary of 397 distinct characters (including special characters and code-mixed characters[8]), and having trained for 31 epochs monitored upon validation loss convergence with a patience of 7 epochs, the model resulted in a perplexity score of 5.18 over the test set. The weights of the character embedding layer were then extracted to be used as pre-trained yet trainable character embeddings for our transduction models. We refer to this method of pre-training as **LM-LSTM-Embed**. The embedding weights are allowed to be trained along with the hyperparameters of the models as freezing these lead to a decrease in performance – a plausible reason being the orthographic and grammatical distinction of Hindi (on which the character embeddings are based) from Bhojpuri (the target language whose orthography and grammar must be reflected by modifying the pre-trained embedding weights).

---

[8] Characters borrowed from other languages.

### 4.5.2 Creating embeddings by averaging fastText embeddings

While using the LM-LSTM-Embed embeddings gives the best performance, we also introduce a novel pre-trained character-level embeddings set for the Devanagari script; these are derived from the 300-D Hindi fastText embeddings (Bojanowski *et al.* 2017). We choose the fastText embeddings over other benchmark embeddings because of their inherent sub-word information preserving property that arises from representing each word as a bag of character *n*-grams[9] and not as an atomic token itself. As has been observed by (Bojanowski *et al.* 2017), such subword-level representation of words is highly useful in capturing morphological structure of words. We start with the existing word embeddings for Hindi.[10] These fastText embeddings have been constructed for each word by averaging the embeddings of the character n-grams that make up the word. We derive the pre-trained embedding of a character pre-trainings by averaging over all word vectors of words in which the character occurs, weighted by the number of times it occurs in each word. We refer to this method of pre-training as **fT-Avg-Embed**. We outline the reason to support this method.

For character n-grams, in the case when n = 1 (and loosely extending for n > 1), each existing word embedding would have been the average of all its characters' embeddings (or when n > 1, groups of characters' embeddings). An approximation of 're-obtaining' the character embeddings from the word embeddings would then be to average over all word vectors of words in which the character occurs. This linear transformation would preserve the 'component' of the embedding of the particular character, as the components of all the other characters would cancel each other out. This justification also takes into the account the fact that the components of all characters that are not 'close' (in the sense of co-occuring in the same context, i.e, word) to the particular character would cancel out each other since their vectors can be thought of as being spatially randomly distributed with respect to the vector of the character in question. This also implies that the resulting character embedding will have contributions from the components of characters that are 'close' to it. We propose this method of computing character embeddings as:

(a) it allows us to circumvent the need for extensive lexical and computational resources required for training character embeddings on a large Hindi monolingual corpus from scratch (while making small compromises on overall downstream accuracy);

(b) it uses pre-existing vectors that have been successfully tested upon a range of tasks (Chaudhary *et al.* 2018), lending the method of deriving character embeddings (from word embeddings) in this manner to any pre-trained word vectors that have been themselves composed from sub-word representations; and

(c) it provides scope to study the notion of character embeddings, vis-a-vis word embeddings, since the semantic notion of a word and its embedding is well understood but the notion of a dense representation of a character is not fully understood yet.

---

[9] Consequently, we observe that using the fastText derived embeddings provide highly consistent results in comparison to using GloVe (Pennington *et al.* 2014) or word2vec (Mikolov *et al.* 2013) character-adaptations over monolingual Hindi corpora.

[10] https://github.com/facebookresearch/fastText/blob/master/docs/pretrained-vectors.md

It is important to note that pre-training using the LM-LSTM-Embed method results in the best transduction performance, and also that using the fT-Avg-Embed pre-trainings does significantly better than random initializations, while incurring lesser computation costs than LM-LSTM-Embed. These differences are shown in Table 3 for Hindi – Bhojpuri and in Table 5 for Hindi – Bangla (additional experiments on this language pair will be described in Section 5.3).

## 5  Results

### 5.1  *Comparing our models to the state-of-the-art model*

Table 2 compares the performances of the best AM and HAN models (after the hyperparameter search) with those of the standard encoder-decoder model, the TN model and the current state-of-the-art (SOTA) model (Sharma and Singh 2017). The BLEU scores of AM, HAN and TN are higher than that of the phoneme-chunk based model used by Sharma and Singh (2017). While TN outperforms the SOTA model in terms of word accuracy (WA), AM and HAN lag behind. The simple

seq2seq model performs the worst among all our models, and fails to match up to SOTA.

TN performs the best due to two main reasons:

1. Residual connections that connect the input character embeddings to the final decoder of the output word: The transduction of a Hindi word can be thought of as making character level edits upon the Hindi word, in the same orthographic space. The residual connections, hence, help in learning these edits by conditioning the final decoder not only on the attention-based representation of the input word but also directly the input word itself.
2. The multi-head attention mechanism of TN helps to model the dependencies of characters in the input and output words, regardless of their distances from each other. This would otherwise have to be learnt from a restricted fixed-sized representation, which is usually an LSTM.

The simple seq2seq model performs the poorest because, perhaps, the generic architecture is not adapted to the task in any manner, based on knowledge about the linguistic properties between cognate-pairs. This simple architecture fails to capture long-range dependencies for 1) longer words, and 2) words in which the dependencies between orthographic segments in the source and target word are not very obviously aligned.

The performance improvements of HAN and AM could be attributed to their attention mechanisms that facilitate better capturing of the intricacies of phonetic and orthographic dependencies of cognates. It is interesting to see that HAN performs worse than the AM, and this might be because of the over-parametrization of HAN. HAN is over-parameterized since it was originally proposed for building a hierarchy over documents where a layer of attention is effected across words in sentences of the document, and another layer is effected across characters in words of sentences. In our case we only deal with individual words, and not documents. This, combined with the fact that we have a small training set, perhaps causes HAN to overfit to the training data and perform poorly on the test set. Consequently, we observe a common trend of recurring last character across the transductions generated by HAN and thus, employ a post-processing step to remove all but the first recurring character appended to these words.

Table 2: Comparison of evaluation metrics among encoder-decoder models: without attention (seq2seq); alignment model (AM); hierarchical attention network (HAN); transformer network (TN); and the phoneme chunk based word transduction model (SOTA)

| Metric | seq2seq | AM | HAN | TN | SOTA |
|--------|---------|-------|-------|--------|-------|
| BLEU | 52.89 | 89.71 | 85.94 | **90.89** | 79.82 |
| SS | 57.22 | 88.03 | 84.05 | **90.23** | – |
| WA | 16.32% | 67.22% | 59.77% | **75.71%** | 64.41% |

This step will be referred to in later sections as the 'post-processing' step.

An elaborate account of errors made on different word-pair types by each of these models is presented in Section 6. These differences in errors occur due to the differences in the models as expounded above.

5.2 *Gains over the state of the art*

Table 2 depicts the performance of the aforementioned models using both the pre-trained character embeddings obtained using averaging (AVG) and that built from the language model (LM). It is evident that TN performs best across all accuracy metrics. This is our best model and using this model achieves gains over SOTA of 11.07 BLEU score points (a percentage gain of 13.9%) and a Word Accuracy gain of 11.3% (a percentage gain of 17.5%). The SOTA paper did not provide information of SS scores, and hence we have not made this comparison.

Table 3 depicts the performance of the AM and HAN model with their embedding layer weights initialized with vectors obtained by averaging from FastText embeddings (ft-Avg), those from language model (LM-LSTM), as well as random and zero initialization.

5.3 *Additional experiments on Hindi - Bangla cognate pairs*

We extend our experiments to transducing from Hindi to Bangla by training on a corpus of Hindi – Bangla cognate pairs comprised of 3220 word pairs. We carried out the same 3:1 split upon this corpus to hold out the test set while making a further split of 0.1 upon the train set (2415 word pairs) to obtain a validation set. The results of the Hind – Bangla experiments are presented in Table 2.

| Initialization methods | AM | | | HAN | | |
|---|---|---|---|---|---|---|
| | BLEU | SS | ep | BLEU | SS | ep |
| ft-Avg-Embed | **87.32** | **85.56** | 59 | **83.14** | **80.89** | 20 |
| LM-LSTM-Embed | 89.71 | 88.03 | 22 | 85.94 | 84.05 | 17 |
| Random | 73.51 | 71.68 | 16 | 64.11 | 61.33 | 12 |
| Zero | 68.44 | 63.10 | 14 | 60.83 | 59.20 | 13 |

Table 3: Effect of pre-trained character embeddings: hyper-parameters are kept the same for both models; *ep* indicates the number of epochs until convergence

| Metric | seq2seq | AM | HAN | TN |
|---|---|---|---|---|
| BLEU | 41.88 | 78.49 | 73.65 | **83.76** |
| SS | 55.12 | 77.10 | 70.22 | **82.59** |
| WA | 9.87% | 59.27% | 47.19% | **71.11**% |

Table 4: Comparison of evaluation metrics between seq2seq, AM, HAN, and TN for Hindi - Bangla cognate pairs.

| Initialization methods | AM | | | HAN | | |
|---|---|---|---|---|---|---|
| | BLEU | SS | ep | BLEU | SS | ep |
| ft-Avg-Embed | 78.49 | 77.10 | 56 | 73.65 | 70.22 | 26 |
| LM-LSTM-Embed | **81.03** | **79.72** | 33 | **75.28** | **73.15** | 24 |
| Random | 72.66 | 72.25 | 19 | 60.07 | 58.46 | 10 |
| Zero | 70.39 | 69.88 | 19 | 59.21 | 55.33 | 12 |

Table 5: Effect of pre-trained character embeddings for Hindi - Bangla cognate pairs

The decline in the scores of all four models across the metrics in Table 4 (compared to Table 2) could be boiled down to two major reasons:

1. *Word formation methods*: Bangla has its roots from the Prakrit or middle Indo-Aryan language, which in turn descended from Sanskrit or old Indo-Aryan language. Hindi also shares roots with Sanskrit. Bangla is therefore, neither a dialect nor an immediate sibling of Hindi. This is unlike Bhojpuri. This also means that Bangla, which has its own Brahmi-derived script, namely, the Bangla script, has possible word formation rules that are quite contrary to Hindi. One such instance is the Bangla consonant clustering mechanism. For example, the noun Vishnu, written as [viRNu] *(an Indian name)* in Hindi, has the consonant cluster *ṣa* + *ṇa* [RN]. While the Hindi consonant cluster ( + ) can easily be decomposed into its constituent letters, the Bangla cluster can form a new letter in itself.

2. *Smaller size of the training corpora*: The training corpora for Hindi – Bangla cognate pairs comprises 2415 word pairs, i.e., 1000 instances less than that of Hindi – Bhojpuri train set. The aforementioned grammatical restrictions shared by Hindi and Bangla make it very demanding to discover more such cognate pairs between the two languages, thus constraining the training corpora for our experiments.

As in Section 5.2, Table 5 depicts the performance of the AM and HAN model with different embedding initializations. We observe that the performance improvements across the metrics alongside the increase in training epochs before convergence of the models remain in line with the trends in Table 3, i.e, LM-LSTM-Embed is a better pre-training strategy than ft-Avg-Embed.

## 6 Error analysis

We analyse the outputs of each model to study a pattern in the most common errors made by each of them. We identify six types of orthographic and lexical errors, and four types of errors related to overall translation quality for a word. While orthographic errors are motivated with respect to the types of graphemes generated by character patterns, quality-related errors focus on overall aspects of the transduction being close to the correct translation. Further, we have been able to make some preliminary correlations between the model architectures (AM, HAN, TN) and the errors they make. The main weakness of the AM model is that since it is predominantly bidirectional LSTM-based and only weakly attention-based as compared to HAN and TN, it tends to bias character predictions towards either the early characters or the later characters in the input sequence, sometimes giving poor results towards the mid-sections. Since it processes input in a sequential manner, it also tends to lose out some orthographic information in the process. The TN model's weakness lies in the fact that it is infamously bad at performing copy mechanisms (Dehghani *et al.* 2018), and hence it fails in places where characters / character-groups have to be preserved in the transduction. The HAN model's trade-off between the LSTM's influence and the attention weights' influence lies between the AM and the TN, and the behaviour it shows with respect to errors it leads to, reflects this is in certain ways. However, these are only approximate inferences that we make retrospectively, with the actual behaviour varying on a case-by-case basis.

| Hindi | Bhojpuri (correct) | AM | HAN | TN |
|---|---|---|---|---|
| [praWA] *(tradition)* | [paraWA] | [pararA] | [paraWA] | [paraWA] |
| [paviwrawA] *(purity)* | [paviwrawA] | [paviwwA] | [paviwraw] | [paviwrawA] |
| [pawrakAroM] *(journalists)* | [pawrakArana] | [pawkAranana] | [pawrA] | [pawrakArana] |

Table 6: Handling halant

### 6.1 Orthographic and lexical errors

#### 6.1.1 Halant

Halants refer to diacritics used to signify the lack of an inherent vowel in written Devanagari scripts. In Devanagari, the halant is represented by a diacritic below the consonant it applies on (e.g., ), while it is represented by not using any vowel after the consonant it applies on in the WX notation (e.g., is represented as simply 'x'). In most of the cases, halants are preserved during translation of Hindi to Bhojpuri words, except a few (e.g. [praWA] *(tradition)* becomes [paraWA]). We study the capability of each model to handle the translation of halants (see Table 6). While HAN and TN perform reasonably well at translating halants, AM tends to replace the character possessing halant with some ligature combined with a neighboring character, e.g. [wka] in [pawkAranana] and [wwa] in [paviwwA]. The possible reason for this could be the better attention capabilities of the HAN and the TN, which helps these models force the halant to be appended in the right place, thus ensuring that the immediate neighboring character of the halant in Hindi is joined with the previous character in the Bhojpuri transduction. The AM, having only limited attention influence in comparison, skips the immediate neighboring character and combines the halant with a later character as the LSTM layer in the AM has seen the later character more recently.

| | Hindi | Bhojpuri (correct) | AM | HAN | TN |
|---|---|---|---|---|---|
| Table 7: Handling vowels | [AnA] *(to come)* | [AilA] | [ayanA] | [AnA] | [ayana] |
| | [GumAnA] *(to stir)* | [GumaAvala] | [GuhAvala] | [GumaAvala] | [GomaAvala] |
| | [mOsI] *(maternal aunt)* | [mausI] | [mausI] | [mausI] | [mausI] |

#### 6.1.2 Handling vowels

Vowels play an important role in the translation of Hindi words to their closely related languages. Our investigations show that TN performs the worst in recognising appropriate vowel translations for vowels used in Hindi words. While the outputs of HAN are the most reasonable ones after the post-processing step (described in section 5.1) of removing repeating characters at the end of the word, the AM model performs moderately well in learning correct vowel translations. The reason for this can be attributed to the fact that a Hindi word's vowels are mostly retained in its Bhojpuri cognate. The TN perhaps performs the worst at retaining vowels as it is infamous for being bad at copy mechanisms (Dehghani *et al.* 2018). These examples are shown in Table 7.

#### 6.1.3 Handling anusvāra

An anusvāra is a diacritic used in a variety of written Indic scripts to denote a nasal sound, either a nasalized vowel or a nasal consonant that is not followed by a vowel. Anusvāra is used often in Hindi, but is absent in Bhojpuri writing. In Devanagari, the anusvāra is represented by a dot () above a character, while it is represented by the letter 'M' in the WX notation. The two general patterns for translation of anusvāra are:

1. replacing them by adding an extra ' [na]' in front of the consonant, e.g. [leKoM] *(writings)*-> [leKana]
2. completely removing them, e.g. [BeMta] *(offering)*-> [Beta]

Our study shows that all three models get confused at choosing the correct rule and generally end up choosing the former one. Table 8) shows a few examples of this. Comparatively, the AM based model performs better than the other two.

| Hindi | Bhojpuri (correct) | AM | HAN | TN |
|---|---|---|---|---|
| [kaTinAyioM]<br>*(hardships)* | [kaTinAyiana] | [kaTinAyiana] | [kaTinAyiana] | [kaTinAyiana] |
| [leKoM]<br>*(writings)* | [leKana] | [leKana] | [leKana] | [leKana] |
| [BeMta]<br>*(offering)* | [Beta] | [Benta] | [Ben] | [Benta] |

Table 8: Handling anusvāra

| Hindi | Bhojpuri (correct) | AM | HAN | TN |
|---|---|---|---|---|
| [barEkRA]<br>*(village's name)* | [baraikRA] | [barabakDa] | [baraikRA] | [baraikRA] |
| [avEjFAnika]<br>*(unscientific)* | [abEjFAnika] | [avEjajAnika] | [abEjianika] | [avEjFAnika] |
| [vqkRa]<br>*(plant)* | [biriCa] | [bakC] | [bikRiRa] | [biriCa] |

Table 9: Handling conjuncts for [kRa] and [jFa]

### 6.1.4 Handling conjuncts for [kRa] and [jFa]

Conjuncts are formed when successive consonants, lacking a vowel in between them, physically join together. The conjuncts for [kRa] and [jFa] are special cases in that they are not clearly derived from the letters making up their components, i.e., the conjunct for [kRa] is [k] + [Ra] and for [jFa] it is [j] + [Fa]. The rules for translation of such conjuncts from Hindi to Bhojpuri are difficult to model (e.g. in the translation [vqkRa] *(plant)* to [biriCa], [kRa] becomes [Ca], while it remains as [kRa] in other cases); and therefore we explore the capability of the models in learning such translations. Our study shows that while the translation for [kRa] is easily learned by the models in most cases, the translation for [jFa] often results in ambiguity. Overall, TN performs the best in learning such translations while AM performs the worst. This, and many of the other errors mentioned in this section, are due to the nature of the writing system used. Table 9 shows a few examples of how different models handle conjuncts.

### 6.1.5 Handling the ā-diacritic/'reph' of [ra]: [rva], [rvA], [rspa]

[ra] has a special diacritic that takes the form of a curved upward dash above the preceding consonants. While translating the Hindi words containing such diacritics to their Bhojpuri counterparts, the diacritic is either replaced completely by or simply kept unchanged. Our re-

Table 10: Handling the ā-diacritic of [ra]

| Hindi | Bhojpuri (correct) | AM | HAN | TN |
|---|---|---|---|---|
| [ahiMsApUrvaka] *(nonviolently)* | [ahinsApUrvaka] | [ahinsApUrvan] | [ahinsApUrasaka] | [ahinsApravaka] |
| [nAcapArtI] *(dance party)* | [nAcpaAratI] | [nAcapArta] | [nAcapArapa] | [nAcapArI] |
| [ParnIcara] *(furniture)* | [ParanIcara] | [ParIcara] | [Pari] | [ParanIcara] |

Table 11: Handling hrasva and dirgh varna

| Hindi | Bhojpuri (correct) | AM | HAN | TN |
|---|---|---|---|---|
| [hima] *(snow)* | [hima] | [hima] | [hima] | [hIma] |
| [dUsrA] *(second)* | [dusar] | [dUsar] | [dUsar] | [dosar] |
| [GI] *(ghee)* | [GIva] | [GI] | [GI] | [GI] |

sults show that, in most cases, the AM model learns to preserve the diacritic as is, while the HAN and the TN models generally replace it by a complete [ra]. 10 shows a few examples of how different models handle these diacritics.

6.1.6  Handling hrasva and deergha varna (short and long vowels)

The Bhojpuri transliterations of most Hindi words show the same long and short vowels appearing after the corresponding consonants (words such as [xUsarA] *(second)* can be treated as exceptions). We observe that for words preserving the nature of vowels (long or short), the AM and HAN based models perform better than that of the TN model while all the models fail to learn the cases where the nature of vowels (long or short) is switched upon translation. Interestingly, the TN model actually recognizes words in which the long vowel has to be switched to the short vowel and vice versa (we deduce this because it does not preserve the long/short nature in these cases) but it does not perform the switch correctly, and instead shows ambiguous behaviour on predictions. Table 11 shows a few examples of how different models handle short and long vowels.

6.2  *Transduction-based properties*

6.2.1  Performance on long words

We consider words exceeding six characters in length to be long words. We found that while the translation quality of each model degrades as word length increases, the AM based architecture is able to maintain the most sensible outputs, followed by the TN and the HAN based models. Table 12 shows examples of the behaviour of different models when they encounter long words.

| Hindi | Bhojpuri (correct) | AM | HAN | TN |
|---|---|---|---|---|
| [vidyArWiyoM]<br>*(students)* | [vidyAraWiyana] | [vidyArWayana] | [vidyArasiiyana] | [vidyAriyana] |
| [surakRAkarmIyoM]<br>*(guards)* | [sorakRAkaramiyana] | [surakFAkararanana] | [surakRAkairIyana] | [surakRArana] |
| [haWiyArabanda]<br>*(armed)* | [haWiyAraband] | [haWiyArabanda] | [haWiyArabanda] | [haWiyArabda] |

Table 12: Performance on long words

| Hindi | Bhojpuri (correct) | AM | HAN | TN |
|---|---|---|---|---|
| [mOlavI]<br>*(Muslim doctor)* | [mOlavI] | [maubi] | [maulavI] | [maulavI] |
| [Jata]<br>*(instant)* | [Jata] | [JaCata] | [Jatatatata], [Jata] | [Jata] |
| [haWiyArabanda]<br>*(armed)* | [haWiyArabanda] | [haWiyArabanda] | [haWiyArabanda] | [haWiyArabda] |

Table 13: Performance on identical translations

### 6.2.2 Identical transduction

For words in Hindi that are written identically in Bhojpuri, we observe that the HAN based model gives the best results after post-processing (described in section 5.1). The TN model fails in cases of longer words while the performance of the AM based model deteriorates for shorter words as well as vowels. Examples of behavior in cases of identical transductions are presented in Table 13.

### 6.2.3 Sensible yet erroneous translations

We individually study the translations made by each model which sound legitimate when compared to the translations of other similar words but are actually wrong. For example, for diphthongs, while the Bhojpuri translation for [BEyA] is [BaiyA] *(elder brother)* ( [BE] replaced by [Bai]), the translation for [kEmarA] *(camera)* does not follow such approach, whereby the character ' [kE]' remains preserved. We observe that each model has its own types of erroneous translations due to such ambiguities as shown in Table 14.

### 6.2.4 Phonetically/orthographically invalid translations

These are predicted transductions which do not follow the necessary phonetic rules in order to be pronounced. We study the type of such words individually for each model. Our findings suggest that the TN performs excellently in learning such rules since we did not notice any such instance of unpronounceable words present in the outputs of TN.

Table 14: Some sensible yet erroneous translations

| Model | Hindi | Bhojpuri (correct) | Predicted |
|---|---|---|---|
| AM | [wanmayawA] *(concentration)* | [wanmayawA] | [wanamayawA] |
| AM | [kEmerA] *(camera)* | [kEmerA] | [kaimerA] |
| HAN | [warabUjA] *(watermelon)* | [warabUjA] | [warabUja] |
| HAN | [yamunA] *(yamuna)* | [yamunA] | [jamun] |
| TN | [KilAdI] *(player)* | [KelAdI] | [KelAdI] |
| TN | [upkaraNa] *(equipment)* | [upakaraNa] | [opkaraNa] |

Table 15: Phonetically/orthographically invalid translations: for invalid predictions, a closest approximation of the WX notation is given

| Model | Hindi | Bhojpuri (correct) | Predicted |
|---|---|---|---|
| AM | [XarmanwaraNa] *(religious conversion)* | [Xaramanwarana] | [Xarmaani] |
| AM | [surakRAkarmIyoM] *(security guards)* | [surakRAkaramIyana] | [surakRAkairIyana] |
| AM | [varjiwa] *(contraband)* | [varajiwa] | [baraimi] |
| HAN | [kuMjI] *(key)* | [kunjI] | [kuMI] |
| HAN | [avEjFAnika] *(unscientific)* | [abEjFAnika] | [avEjajAnika] |

Overall, HAN produced the most unpronounceable translations as can be seen from Table 15.

### 6.3  *Limitations of the method*

While training the system on cognate pairs ensures that the model does not require exhaustive resources such as a parallel corpora, the fact remains that cognate-property is inherently confined to the case of closely-related languages. As described in Section 1, Bhojpuri is closely related to Hindi; it is a dialect/immediate sibling and shares the same script.[11] By contrast, Bangla (section 5.3) is a direct descendant of Sanskrit or Pali[12] to Prakrit and Apabhramsha (Chatterji (1926); Bali (2016)) but is only loosely related to Hindi: it is neither a dialect nor an immediate sibling of Hindi; it has its own script; and it has word formation rules that are sometimes contrary to Hindi. The effects of such increased 'language distance' are apparent from Table 2 (Hindi – Bhojpuri) and Table 4 (Hindi – Bangla) where Hindi – Bhojpuri cognate pairs show far better transduction quality than Hindi – Bangla.

---

[11] Although Bhojpuri was historically written in Kaithi scripts, those have gone obsolete with Devanagari replacing as the primary script.

[12] Hindi descended from Prakrit, Apabhramsha and Perso-Arabic while Bangla and Bhojpuri inherited the regional dialects of Apabhramsha. Consequently, Bangla and Bhojpuri, sharing several regional boundaries of Northern India, Bangladesh and Nepal, can be thought of having similar ancestral languages to Hindi.

Further, with the increased distance between the source and the target language in a language pair, it is safe to say that the set of cognates shared by these dwindles (Beel and Felder 2014), since cognates, by definition, are word pairs that not only have similar meaning and phonetics but also reflect an allied linguistic derivation. We observed this even in our own cognate datasets: the Hindi – Bangla dataset had 25% fewer cognates than the Hindi – Bhojpuri dataset (see Section 5.3). Having fewer real-world cognate pairs means that even if a nominally complete dataset of all cognates were to be compiled, there would still be comparatively less data to train models on, hence exhibiting another limitation in extending the method to language pairs of arbitrary distance.

## 7   Improvements on Hindi - Bhojpuri machine translation

This section talks of the improvements we make on machine translation from Hindi to Bhojpuri. The authors would like to mention two key points here. First, this section stands distinguished from the rest of the paper in that the methods and results discussed thus far hold for word-to-word transduction. In contrast, we now depict how such transductions can be used to improve the accuracy of a machine translation system. Second, we must mention that to the best of our knowledge, no prior machine translation systems have been trained on a Hindi – Bhojpuri parallel corpus, simply because such a corpus does not exist.

For our purposes, we build an artifical parallel corpus in the following manner. We first scrape four Bhojpuri blogs for Bhojpuri sentences : Anjoria[13], TatkaKhabar[14], Bhojpuri Manthan[15] and Bho-jpuri Sahitya Sarita[16]. From these, we collect a set of approximately 40 000 Bhojpuri sentences. We then follow a two-step process for obtaining Hindi translations of these sentences using the Google Translate API. Since the API does not offer Bhojpuri as a source language, we first use the Hindi-English API for translating the Bhojpuri sentences to English (as an intermediate language). We then translate these English sentences to Hindi using the English-Hindi API. Although this process gives us noisy translations between Bhojpuri and Hindi, this method is based upon the hypotheses that:

1. this is the best available solution in the absence of Hindi – Bhojpuri parallel corpora, let alone a pre-trained Hindi – Bhojpuri translation system; and
2. Bhojpuri is sufficiently similar to Hindi that its linguistic properties remain preserved satisfactorily throughout the intermediate processing.

---

[13] https://www.anjoria.com/
[14] http://khabar.anjoria.com/
[15] http://bhojpurimanthan.com/

Table 16: Machine translation corpus statistics

|  | Training set | Test set |
|---|---|---|
| Number of sentences | 40 000 | 1000 |
| Total number of tokens | 812 070 | 19 689 |
| Number of unique tokens | 20 551 | 620 |

Also, the choice of English as an intermediate language is based on the fact that it gives the best BLEU scores when Hindi is fixed as the other language in the translation pair. At the time of writing this work, there is no Bhojpuri-Hindi API (or publicly reported Hindi – Bhojpuri machine translation system), while trivially translating using the Hindi-Hindi API does not seem to work as the API simply copies the inputs to outputs. For reporting the machine translation results, we use the standard document-level BLEU score as suggested by Papineni *et al.* (2002).

For the held-out test set, we curated 1000 sentences for which ground truth translations were manually obtained from experts, and not artificially generated. Statistics on training and test sentences can be found in Table 16.

We train a Bi-LSTM based encoder-decoder network (500 units) with Luong attention, as described in Luong *et al.* (2015a) (we use OpenNMT's global attention implementation) on the training set for

---

[16] http://www.bhojpurisahityasarita.com/

Hindi – Bhojpuri translation. The model thus trained resulted in a BLEU score of around 7.1 on the test set, which is not surprising given the method of training. With this as the baseline, we make corrections to this model's output. We first align each Hindi sentence and its Bhojpuri translation, by aligning pairs of source-target words based on their pairwise weights in the attention matrix. Following this, we identify OOV Hindi words using a simple dictionary-based approach as has been suggested by Bahdanau *et al.* (2014). We use a shortlist of 15 000 most common words in Hindi, obtained from the Hindi Wikipedia monolingual corpus, and treat **all other words** as OOV. We experiment with 5 000, 10 000, 15 000, and 20 000 most common words, and find the 15 000 word shortlist produced the best BLEU score improvements. We then replace the translation (as obtained by the alignments using the attention matrix) of each OOV word with its corresponding transduction generated by our word transduction model. Replacing the translation of OOV words with that of their transductions leads to an improvement of 6.3 points in the BLEU score, which is substantial considering that we are translating to a low-resource language. We obtain a BLEU score of 13.4 with such a basic translation set-up followed by simple correction of OOV translations using transductions.

More importantly, the improvement in the MT BLEU score shows that even though the task of transduction is a focused one, it generalizes well to OOV Hindi words that are not part of a Hindi – Bhojpuri cognate pair. This stands as an important aspect of our work. Our transduction model, despite being trained on a dataset that is constrained to cognate pairs, lends reasonable improvements to the highly generalized machine translation task by exploiting the closeness of Hindi and Bhojpuri.

## 8    Discussion

We propose a character-level transduction of OOV words between a pair of closely related languages, out of which at least one is a low-resource language. Word transduction aims to predict the orthographic form of the word in the target language, given the word in the source language. We restrict the training space to a set of cognates, since in the case of closely-related languages, a cognate can be a good approximation to a translation, if not the translation itself. We present three different models for the same, each of which performs well on handling certain types of grapheme transformations, while performing sub-optimally on others. While all our models[17] outperform the current state of art for Hindi – Bhojpuri transduction, the TN model gives the overall best performance. We suggest a two-step procedure to improve a low-resource NMT system by:

1. identifying the need to handle OOV words separately; and
2. transducing them to their target equivalents, instead of translating.

In the process, we also propose a primitive yet useful MT method using Google Translate APIs for a pair of languages which has no known MT system. In the future, we would like to test our models on more closely related language pairs. Further, we would like to build a state-of-art MT pipeline for low resource languages, which incorporates our method to handle OOV words. We would also like to compare the effect of transduction as a solution to the OOV problem faced by word-level MT systems by comparing it to a character-level MT system.

---

[17] We open source the code and data curated for this work in our github repo: https://github.com/Saurav0074/nmt-based-word-transduction.

*Jha, Sudhakar, and Singh*Naihan LI, Shujie LIU, Yanqing LIU, Sheng ZHAO, Ming LIU, and Ming ZHOU (2018), Close to human quality TTS with transformer, *arXiv preprint arXiv:1809.08895*.

Minh-Thang LUONG and Christopher D. MANNING (2016), Achieving Open Vocabulary Neural Machine Translation with Hybrid Word-Character Models, *Computing Research Repository (CoRR)*, abs/1604.00788.

Thang LUONG, Hieu PHAM, and Christopher D. MANNING (2015a), Effective Approaches to Attention-based Neural Machine Translation, in *Proceedings of the Conference on Empirical Methods in Natural Language Processing (EMNLP)*, pp. 1412–1421, Lisbon, Portugal.

Thang LUONG, Ilya SUTSKEVER, Quoc V. LE, Oriol VINYALS, and Wojciech ZAREMBA (2015b), Addressing the Rare Word Problem in Neural Machine Translation, in *Joint conference of the 47th Annual Meeting of the Association for Computational Linguistics and the 4th International Joint Conference on Natural Language Processing of the Asian Federation of Natural Language Processing*, pp. 11–19, Beijing, China.

B.P. MAHAPATRA, P. PADMANABHA, G.D. MCCONNELL, and V.S. VERMA (1989), *The Written Languages of the World: A Survey of the Degree and Modes of Use : Volume 2: India: Book 1 Constitutional Languages*, volume 1, Quebec : Presses de l'Université Laval; Forest Grove, Oregon.

Gideon S MANN and David YAROWSKY (2001), Multipath translation lexicon induction via bridge languages, in *Proceedings of the Second Meeting of the North American Chapter of the Association for Computational Linguistics on Language technologies*, pp. 1–8, Association for Computational Linguistics, Pittsburgh, PA, USA.

Tomas MIKOLOV, Ilya SUTSKEVER, Kai CHEN, Greg S CORRADO, and Jeff DEAN (2013), Distributed representations of words and phrases and their compositionality, in *Advances in Neural Information Processing Systems (NIPS)*, pp. 3111–3119.

Diwakar MISHRA and Kalika BALI (2011), A Comparative Phonological Study of the Dialects of Hindi, in *Proceedings of the International Congress of Phonetic Sciences (ICPhS)*, pp. 1390–1393, Hong Kong.

Preslav NAKOV and Jörg TIEDEMANN (2012), Combining word-level and character-level models for machine translation between closely-related languages, in *Proceedings of the 50th Annual Meeting of the Association for Computational Linguistics: Short Papers-Volume 2*, pp. 301–305, Association for Computational Linguistics (ACL), Jeju Island, Korea.

Yurii NESTEROV (1983), A method of solving a convex programming problem with convergence rate O (1/k2), *Soviet Mathematics Doklady*, 27(2):372–376.

Kishore PAPINENI, Salim ROUKOS, Todd WARD, and Wei-Jing ZHU (2002), BLEU: a method for automatic evaluation of machine translation, in *Proceedings*[ 40 ]

# Appendices

## A     Tuning AM and HAN

### A.1     *Encoding/decoding layer size*

For each experiment, we maintain equal encoding and decoding layer sizes. A good rule of thumb is to begin with half the size of the embedding vector for each token (i.e., 150 dimensional in our case). We therefore experiment by varying the layer sizes in the range 60, ..., 100 with an interval of 10. Our experiments show that a hidden layer size of 80 LSTM cells give the optimum BLEU score after fixing the number of encoder-decoder layers as 1 – 1, batch size as 8, a dropout of 0.2 and the optimizer as Adam (Kingma and Ba 2015) for both the models.

### A.2     *LSTM versus GRU*

For AM, we observe that the difference in BLEU score between using an LSTM and a GRU as the encoding-decoding units is not significant; while the Levenshtein distance increases by almost two points when GRU is used (Table 17). On the other hand, the performance of the HAN based model degrades significantly when using a GRU. Using GRU as the en- coding/decoding unit results in faster convergence of both the models (fewer epochs) because there are fewer weight parameters to be updated at every epoch. Furthermore, our experiments show that LSTMs outperform GRUs in handling translations of longer words for both the models. The lack of controlled exposure of the memory content (Chung *et al.* (2014), Cho *et al.* (2014)) in the GRU cell could explain this behaviour.

Table 17: Comparison of BLSTM and BGRU as encoding/decoding units; all other tuning parameters fixed

| Models | BLSTM | | | BGRU | | |
|---|---|---|---|---|---|---|
| | BLEU | SS | ep | BLEU | SS | ep |
| AM | **86.43** | **81.23** | 26 | **85.94** | **83.31** | 21 |
| HAN | 83.14 | 80.89 | 16 | 76.15 | 73.77 | 16 |

Fixed parameters: number of encoder-decoder layers = 1-1 and 1-2 for AM and HAN respectively, hidden layer size = 80, batch size = 8, dropout = 0.2, and Adam optimizer

Table 18: Comparison of number of layers with fixed batch size of 8

| #encoder layers | #decoder layers | AM | | | HAN | | |
|---|---|---|---|---|---|---|---|
| | | BLEU | SS | ep | BLEU | SS | ep |
| 1 | 1 | **87.34** | **83.98** | 43 | 76.12 | 75.09 | 12 |
| 1 | 2 | 84.12 | 81.73 | 37 | **83.14** | **80.89** | 16 |
| 1 | 3 | 83.47 | 80.01 | 35 | 78.8 | 78.03 | 11 |
| 2 | 1 | 83.88 | 80.96 | 48 | 74.63 | 73.14 | 22 |
| 2 | 2 | 85.12 | 82.20 | 41 | 75.94 | 75.5 | 18 |

Fixed parameters: number of encoder-decoder layers = 1-1 and 1-2 for AM and HAN respectively, hidden layer size = 80, dropout = 0.2, and Adam optimizer with initial learning rate = $10^{-3}$.

### A.3  *Number of layers*

We find that the optimum number of encoder and decoder layers for the AM architecture is 1, while the HAN based model delivers optimum BLEU score (83.14) with the depth of encoder and decoder layers being 1 and 2, respectively (Table 18).

### A.4  *Batching*

Our experiments show that a batch size of eight results in maximum BLEU scores (87.34 and 83.14 respectively) for both the models (Table 19). The improvement in BLEU points while increasing batch size to eight from one can be attributed to the fact that the variance of the stochastic gradient update reduces with the increase in batch size. . However, there seems to be no notable increase in the BLEU score for batch sizes greater than eight. Also, it can be inferred from Table 19 that the SS metric shows no particular trend with the increase in batch size.

| Batch size | AM | | | HAN | | |
|---|---|---|---|---|---|---|
| | BLEU | SS | ep | BLEU | SS | ep |
| 1 | 85.52 | **84.34** | 24 | 74.80 | 73.06 | 10 |
| 4 | 86.43 | 81.23 | 26 | 78.77 | 76.23 | 14 |
| 8 | **87.34** | 83.98 | 43 | **83.14** | **80.89** | 16 |
| 16 | 87.05 | 83.41 | 61 | 79.13 | 78.39 | 20 |
| 20 | 86.81 | 82.65 | 72 | 79.04 | 76.64 | 23 |

Table 19: Comparison of batch sizes

Fixed parameters: number of encoder-decoder layers = 1-1 and 1-2 for AM and HAN respectively, hidden layer size = 80, dropout = 0.2, and Adam optimizer with initial learning rate = $10^{-3}$.

A.5 *Optimization methods*

We tried varying the optimization methods along with their respective learning rates for the AM and HAM models:

- Adam (Kingma and Ba 2014b) optimizer with and without learning rate decay
- Stochastic Gradient Descent (SGD)
- SGD with momentum (mom) (Rumelhart *et al.* (1986); Sutskever *et al.* (2014a); Zinkevich *et al.* (2010))
- SGD with Nesterov momentum (Nesterov (1983))
- RMSprop (Hinton *et al.* (2012a))
- Adadelta (Zeiler (2012))
- Adagrad (Duchi *et al.* (2011))

We concluded that the Adam optimizer with a learning rate of $10^{-3}$ performs the best (see Table 20). While the use of decay with SGD boosts its potential, the performance of Adam deteriorates badly when trained with decay of the initial learning rate.

Out of all the optimizers, Adadelta performs the worst, giving a constant validation accuracy after being trained for more than a certain number (usually < 6) of epochs (denoted by *Model broken* in Table 20). HAN performs worse with RMSprop and Adagrad, and is thus much more sensitive to the optimizer used. SGD with a learning rate of $10^{-3}$ when used with decay and Nesterov momentum gives results that are on par with that of Adam.

Table 20: Comparison of optimization methods

| Method | lr | details | AM BLEU | AM SS | AM ep | HAN BLEU | HAN SS | HAN ep |
|---|---|---|---|---|---|---|---|---|
| Adam | $10^{-2}$ | - | 77.35 | 75.58 | 17 | 67.18 | 64.34 | 13 |
| | $10^{-3}$ | - | 86.43 | 81.23 | 26 | 83.14 | 80.89 | 16 |
| | $10^{-3}$ | decay = $9*10^{-1}$ | 9.68 | 3.20 | 33 | Model broken | | |
| SGD | $0.5*10^{-3}$ | - | 81.57 | 79.86 | 21 | 71.96 | 68.10 | 27 |
| | $10^{-2}$ | decay = $10^{-6}$, Nesterov | 76.9 | 77.99 | 44 | 66.61 | 58.95 | 33 |
| | $0.5*10^{-2}$ | decay = $10^{-6}$, Nesterov | 70.6 | 68.06 | 24 | 65.64 | 57.35 | 94 |
| | $0.5*10^{-2}$ | decay = $10^{-6}$ | 77.29 | 76.28 | 48 | 48.24 | 57.96 | 88 |
| | $10^{-3}$ | decay = $10^{-6}$, Nesterov | 81.92 | 80.02 | 329 | 74.91 | 73.08 | 297 |
| RMSprop | $10^{-3}$ | - | 79.88 | 78.89 | 38 | 60.12 | 58.04 | 16 |
| Adagrad | $10^{-2}$ | - | 75.71 | 77.9 | 386 | 49.05 | 56.52 | 278 |
| | $10^{-3}$ | - | 53.78 | 25.18 | 418 | 36.44 | 43.19 | 463 |
| Adadelta | $5*10^{-1}$ | decay = $9*10^{-1}$ | Model broken | | | 12.52 | 2.07 | 83 |
| | 1.0 | decay = $9*10^{-1}$ | Model broken | | | Model broken | | |
| | 1.0 | decay = $95*10^{-2}$ | Model broken | | | Model broken | | |

Fixed parameters: number of encoder-decoder layers = 1-1 and 1-2 for AM and HAN respectively, hidden layer size = 80, dropout = 0.2, and batch size = 8.

Table 21: Comparison of regularization techniques

| Dropout | l2 | AM BLEU | AM SS | AM ep | HAN BLEU | HAN SS | HAN ep |
|---|---|---|---|---|---|---|---|
| 0 | 0 | 83.8 | 82.64 | 33 | 71.30 | 68.62 | 11 |
| | $10^{-2}$ | 81.80 | 79.64 | 36 | 66.88 | 65.09 | 16 |
| 0.1 | 0 | 81.32 | 80.66 | 33 | 73.81 | 71.67 | 14 |
| 0.2 | 0 | 86.65 | 85.10 | 36 | **83.14** | **80.89** | 16 |
| | $10^{-1}$ | **87.32** | **85.56** | 59 | 75.06 | 73.11 | 14 |
| | $10^{-2}$ | 84.83 | 82.77 | 49 | 75.13 | 73.25 | 21 |
| | $10^{-3}$ | 82.80 | 79.65 | 38 | 72.37 | 71.77 | 19 |
| 0.3 | 0 | 83.22 | 82.68 | 41 | 78.63 | 76.35 | 18 |
| 0.4 | 0 | 82.78 | 81.61 | 47 | 76.94 | 75.18 | 23 |
| 0.5 | 0 | 80.97 | 80.22 | 53 | 73.07 | 72.22 | 31 |
| 0.6 | 0 | 80.03 | 79.55 | 61 | 70.30 | 68.01 | 37 |
| 0.7 | 0 | 78.81 | 77.94 | 69 | 65.86 | 64.21 | 46 |

Fixed parameters: number of encoder-decoder layers = 1-1 and 1-2 for AM and HAN respectively, hidden layer size = 80, dropout = 0.2, and Adam optimizer with initial learning rate = $10^{-3}$.

## A.6 *Regularization methods*

We conduct our experiments on two types of regularization techniques: dropout (Hinton *et al.* (2012b)) and L2 regularization (Hanson and Pratt (1989); Weigend *et al.* (1991)). We study the effects of each technique individually as well as of combining both the techniques. The dropout rate was varied in the range 0 – 0.7. It can be concluded from Table 21 that the performance of the both the models keeps improving until the dropout rate of 0.2 (i.e., turning off 20% of the activations) and start declining on increasing the rate further. The combination of dropout rate of 0.2 and l2 rate of 0.1 delivers the optimum results for AM while the HAN model performs best in absence of l2 regularization.